\documentclass[11pt,a4paper]{article}
\usepackage[hyperref]{emnlp2018}
\usepackage{times}
\usepackage{latexsym}

\usepackage{url}
\usepackage{natbib}
\usepackage{algorithm}
\usepackage{algpseudocode}
\usepackage{amsmath}
\usepackage{amssymb}
\usepackage{amsthm}
\usepackage{bibentry}
\usepackage{booktabs}
\usepackage{tablefootnote}
\usepackage{caption}
\usepackage{comment}
\usepackage{fancyhdr}
\usepackage{float}
\usepackage{graphicx}
\usepackage{hyperref}
\usepackage{latexsym}
\usepackage{multirow}
\usepackage{paralist}
\usepackage{setspace}
\usepackage{subcaption}
\usepackage{tabularx}
\usepackage{tikz}
\usetikzlibrary{calc}

\usepackage{url}
\usepackage{microtype}
\usepackage{xspace}

\newfloat{Algorithm}{tb}{lox}

\usepackage[utf8]{inputenc}
\usepackage[T1]{fontenc}
\usepackage[german,english]{babel}

\DeclareMathOperator*{\argmax}{arg\,max}

\usepackage{amssymb}
\usepackage{booktabs}
\usepackage{tikz}
\usepackage{multirow}
\usepackage{pgfplots}
\usepackage{tikzscale}
\usepackage{caption}
\usepackage{subcaption}
\usepackage{diagbox}
\usepackage{array}
\newcolumntype{L}[1]{>{\raggedright\let\newline\\\arraybackslash\hspace{0pt}}m{#1}}
\newcolumntype{C}[1]{>{\centering\let\newline\\\arraybackslash\hspace{0pt}}m{#1}}
\newcolumntype{R}[1]{>{\raggedleft\let\newline\\\arraybackslash\hspace{0pt}}m{#1}}
\newcommand{\ment}[1]{[{\bf #1}]}
\newcommand{\doc}{\mathcal{D}}

\newcommand{\name}[1]{{\sc Xelms}}
\newcommand{\feedfwd}{\mathcal{F}}
\newcommand{\ignore}[1]{}
\usepackage{CJKutf8}
\usepackage{enumitem}

\newcommand{\todo}[1]{}
\newcommand{\tododr}[1]{}
\aclfinalcopy % Uncomment this line for the final submission
\title{\vspace*{-0.5in}
{\small \hfill {EMNLP'18}}
\\
\vspace*{.25in} Joint Multilingual Supervision for Cross-lingual Entity Linking}
\author{Shyam Upadhyay \\
  University of Pennsylvania \\
  Philadelphia, PA \\
  {\tt shyamupa@seas.upenn.edu}\\\And
  Nitish Gupta \\
  University of Pennsylvania \\
  Philadelphia, PA \\
  {\tt nitishg@seas.upenn.edu} \\\And
  Dan Roth \\
  University of Pennsylvania \\
  Philadelphia, PA \\
  {\tt danroth@seas.upenn.edu}}

\date{}

\begin{document}
\maketitle
\begin{abstract}
  Cross-lingual Entity Linking (XEL) aims to ground entity mentions written in {\em any} language to an English Knowledge Base (KB), such as Wikipedia.
XEL for most languages is challenging, owing to limited availability of resources as supervision. 
We address this challenge by developing
the first XEL approach that combines {supervision} from multiple languages {\em jointly}.
This enables our approach  to:
{\bf (a)} augment the limited supervision in the target language with additional supervision from a high-resource language (like English), and 
{\bf (b)} train a {\em single} entity linking model for
multiple languages, improving upon individually trained models for each language.
Extensive evaluation on three benchmark datasets across 8 languages shows that our approach significantly improves over the current state-of-the-art.
We also provide analyses in two limited resource settings:
{\bf (a)} {\em zero-shot setting}, when no supervision in the target
language is available, and in
{\bf (b)} {\em low-resource setting}, when some supervision in the target language is available. Our analysis provides insights into the limitations of zero-shot XEL approaches in realistic scenarios, and shows the value of joint supervision in low-resource settings.\footnote{Code at \url{www.github.com/shyamupa/xelms}}

%%% Local Variables:
%%% mode: latex
%%% TeX-master: "main"
%%% End:
\end{abstract}

\section{Introduction}
\label{sec:intro}
\begin{figure}
  \centering
  \includegraphics[scale=0.54]{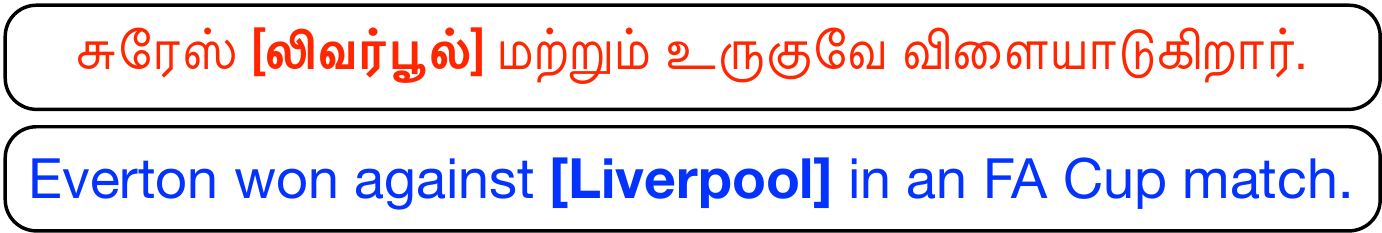}
  \caption{\footnotesize Tamil and English mention contexts containing \ment{mentions} of the entity {\tt Liverpool\_F.C.} from the respective Wikipedias. Tamil Wikipedia only has 9 mentions referring to {\tt Liverpool\_F.C.}, whereas English Wikipedia has 5303 such mentions. Clearly, there is a need to augment the limited contextual evidence in low-resource languages with evidence from high-resource languages like English.
Tamil sentence translates to ``Suarez plays for \ment{Liverpool} and Uruguay.''
  }
  \label{fig:intro}
\end{figure}

Entity Linking (EL) systems ground entity mentions in text to entries in Knowledge Bases (KB), such as
Wikipedia~\cite{MihalceaCs07}. Recently, the task of
Cross-lingual Entity Linking (XEL) has gained attention~\cite{mcnamee2011cross,ji2015overview,TsaiRo16b} with the goal of grounding
entity mentions written in {\em any} language to
the English Wikipedia. 
For instance, Figure~\ref{fig:intro} shows a Tamil (a language with $>$70 million speakers) and an English mention (shown \ment{enclosed}) and their mention contexts. XEL involves grounding the Tamil mention (which translates to `Liverpool') to the football club {\tt Liverpool\_F.C.}, and not the city or the university.
XEL enables knowledge acquisition directly from documents in any language, without resorting to machine translation.

\begin{figure*}[t!]
  \begin{subfigure}{.5\textwidth}
    \raggedright
    \includegraphics[scale=0.45]{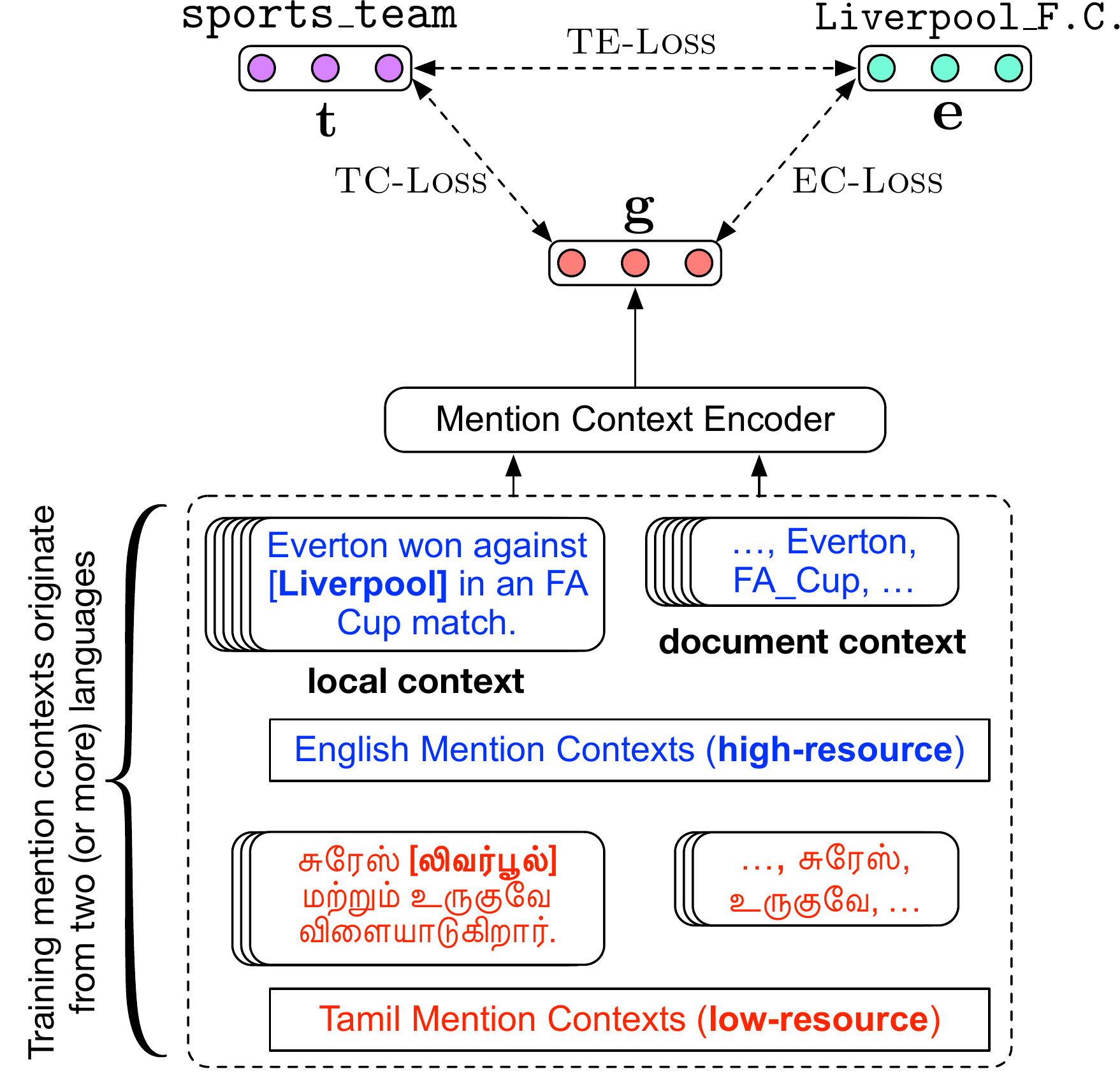}
    \caption{Overview of \name~. Mentions are shown \ment{enclosed}.}
    \label{fig:modelarch}
  \end{subfigure}
  \begin{subfigure}{.5\textwidth}
    \centering
    \raggedleft
    \includegraphics[scale=0.45]{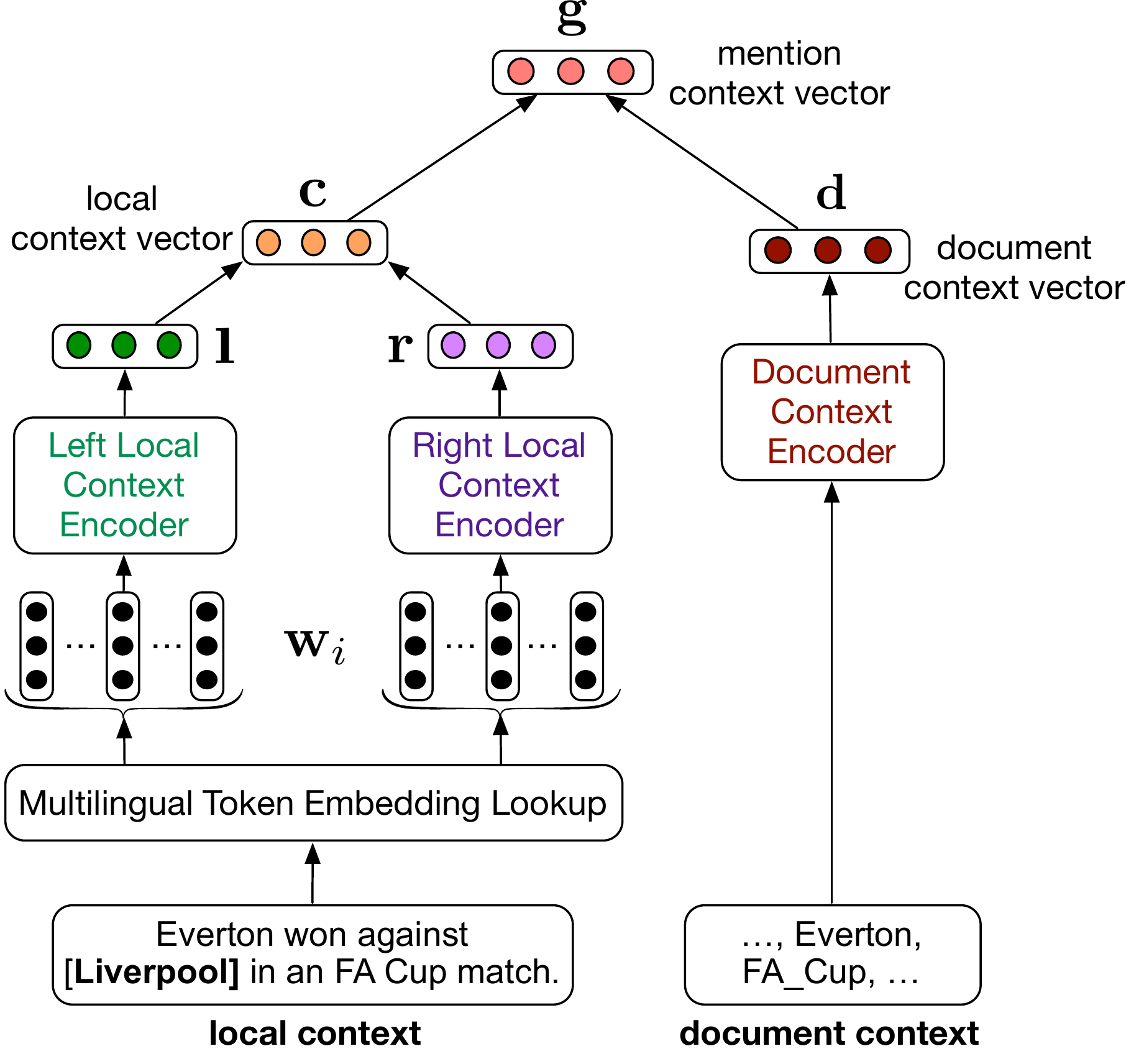}
    \caption{Mention Context Encoder.}
    \label{fig:mentionenc}
  \end{subfigure}
  \caption{\footnotesize
  ({\bf a}) Grounded mentions from two or more languages (English and Tamil shown) can be used to supervise \name~.
  The context $\mathbf{g}$, entity $\mathbf{e}$ and type $\mathbf{t}$ vectors interact through Entity-Context loss ({\sc EC-Loss}), Type-Context loss ({\sc TC-Loss}) and Type-Entity loss ({\sc TE-Loss}).
  The Tamil sentence is the same as in Figure~\ref{fig:intro}, and other mentions in it translate to \ment{Suarez} and \ment{Uruguay}.
  ({\bf b}) The Mention Context Encoder (\S\ref{ssec:encoder}) encodes the local context (neighboring words) and the document context (surfaces of other mentions in the document)  of the mention into $\mathbf{g}$. Internal view of local context encoder is in Figure~\ref{fig:loccxt}.}
  \end{figure*}

Training an EL model requires grounded mentions, i.e. mentions of entities that are grounded to a Knowledge Base (KB), as supervision (Figure~\ref{fig:intro}).
While millions of such mentions are available 
in English, by virtue of hyperlinks in the English Wikipedia, this is not the case for most languages. This makes learning XEL models challenging, especially for languages with limited resources (e.g., the Tamil Wikipedia is only 1\% of the English Wikipedia in size).
To overcome this challenge, it is desirable to 
augment the limited contextual evidence available in the target language with evidence from high-resource languages like English.  

We propose \name~ (\underline{XEL} with \underline{M}ultilingual \underline{S}upervision) (\S\ref{sec:cross-lingual-entity}), the first approach that fulfills the above desiderata by using multilingual supervision to train an XEL model. 
\name~ represents the mention contexts of the same entity from different languages in the same semantic space using a single context encoder (\S\ref{ssec:encoder}). 
Language-agnostic entity representations are
jointly learned with the relevant mention context representations, so that an entity and its context share similar representations. Additionally, by encoding freely available structured
knowledge, like fine-grained entity types, the entity and context representations can be further improved  
(\S\ref{subsec:type}).

The ability to use multilingual supervision enables \name~ to learn XEL models for target languages with limited resources by exploiting freely available supervision from high resource
languages (like English). We show that \name~  outperforms existing state-of-the-art
approaches that only use target language supervision, across 3 benchmark datasets in 8 languages
(\S\ref{sec:main-experiments}). 
Moreover, while previous XEL models~\cite{mcnamee2011cross,TsaiRo16b} train separate models for different languages, \name~ can train a {\em single} model for performing XEL in multiple languages (\S\ref{sec:mult-train}). 

One of the goals of XEL is to enable understanding of languages with limited resources.
We provide experimental analyses in two such settings. In the {\em zero-shot setting} (\S\ref{sec:zero-shot-setting}), where \emph{no} supervision is available in the target language, we show that the good performance of zero-shot XEL approaches~\cite{Sil2018} can be attributed to the use of prior probabilities. 
These probabilities are computed from large amount of grounded mentions, which are not available in realistic zero-shot settings. 
In the {\em low-resource setting} (\S\ref{sec:how-much-target}), where some supervision is available in the target language, we show that even when only a fraction of the available supervision in the target language is provided, \name~ can achieve competitive performance by exploiting supervision from English.

The contributions of our work are,
\begin{itemize}[noitemsep,nolistsep]%\setlength\itemsep{0em}
\item A new XEL approach, \name~, that learns a XEL model for a language with limited resources by exploiting additional supervision from a high-resource language like English.
\item \name~ can also train a {\em single} XEL model for multiple languages jointly, which we show improves on separately trained models.
\item Analysis of XEL approaches in the zero-shot and low-resource settings. Our analysis reveals that in realistic scenarios, zero-shot XEL 
is not as effective as previously shown. 
We also show that in low-resource settings jointly training with English leads to better utilization of  target language supervision.
\end{itemize}

%%% Local Variables:
%%% mode: latex
%%% TeX-master: "main"
%%% End:

\section{Cross-lingual EL with \name~}
\label{sec:cross-lingual-entity}
Given a mention $m$ in a document $\doc$ written in any language, XEL involves linking $m$ to its gold entity $e^*$ in a KB, $\mathcal{K}=\{e_1,\cdots,e_n\}$.

An overview of \name~ is shown in Figure~\ref{fig:modelarch}.
\name~ computes the probability, $\text{P}_{\text{context}}(e \mid m)$, of a mention $m$ referring to entity $e \in \mathcal{K}$ using a mention context vector $\mathbf{g} \in \mathbb{R}^h$ representing $m$'s context, and an entity vector $\mathbf{e} \in \mathbb{R}^h$, representing the entity $e \in \mathcal{K}$ (one vector per entity).
\name~ can also incorporate structured knowledge like
fine-grained entity types (\S\ref{subsec:type}) using a
multi-task learning approach~\cite{caruana1998multitask}, by
learning a type vector $\mathbf{t} \in \mathbb{R}^h$ for each
possible type $t$ (e.g., {\tt sports\_team}) associated with the entity $e$.
The entity vector $\mathbf{e}$, context vector $\mathbf{g}$ and the type vector
$\mathbf{t}$ are jointly trained, and interact through appropriately defined pairwise loss
terms -- an Entity-Context loss ({\sc EC-Loss}), Type-Entity loss ({\sc
  TE-Loss}) and a Type-Context loss ({\sc TC-Loss}).

The mention context vector $\mathbf{g}$ is generated by a
mention context encoder (\S\ref{ssec:encoder}), shown in Figure~\ref{fig:mentionenc}. The {\em mention context} of $m$ in
a document $\doc$ consists of: {\bf (a)} neighboring
words around the mention, which we refer to as its {\em local context} and, {\bf (b)} surfaces of other mentions
appearing in $\doc$, which we refer as its {\em document context}.

\noindent \name~ is trained using grounded mentions
 in multiple languages (English and Tamil in
Figure~\ref{fig:modelarch}), which can be derived from Wikipedia (\S\ref{sec:training-data}). 

\subsection{Mention Context Representation}
\label{ssec:encoder}
To learn from mention contexts in multiple languages, we
generate mention context representations using
a language-agnostic mention context encoder. An overview of the mention context encoder is shown in Figure~\ref{fig:mentionenc}.
Below we describe the components of the mention context encoder, namely multilingual word embeddings and local and document context encoders.

\paragraph{Multilingual Word Embeddings}
\cite{ammar2016massively,smith2017offline, Duong-EtAl:2017:EACLlong} jointly encode words in multiple ($\ge$2)
languages in the same vector space such that semantically similar
words in the same language, and translationally equivalent words in
different languages are close (per cosine similarity). Multilingual embeddings generalize
bilingual embeddings, which do the same for two languages {\em only}.

We use {\sc FastText}~\cite{TACL999,smith2017offline}, which aligns monolingual embeddings of multiple languages in the same space using a small dictionary ($\sim$2500 pairs) from each language to English. Both monolingual embeddings and the dictionary can be easily obtained for languages with limited resources.
We denote the multilingual word embeddings for a
set of tokens $\{w_1,w_2,\cdots,w_n\}$ by
$\mathbf{w}_{1:n} =\{\mathbf{w}_1,\mathbf{w}_2,\cdots,\mathbf{w}_n\}$, where each
$\mathbf{w}_i \in \mathbb{R}^d$.

\begin{figure}[t]
  \footnotesize
  \centering
\includegraphics[scale=0.45]{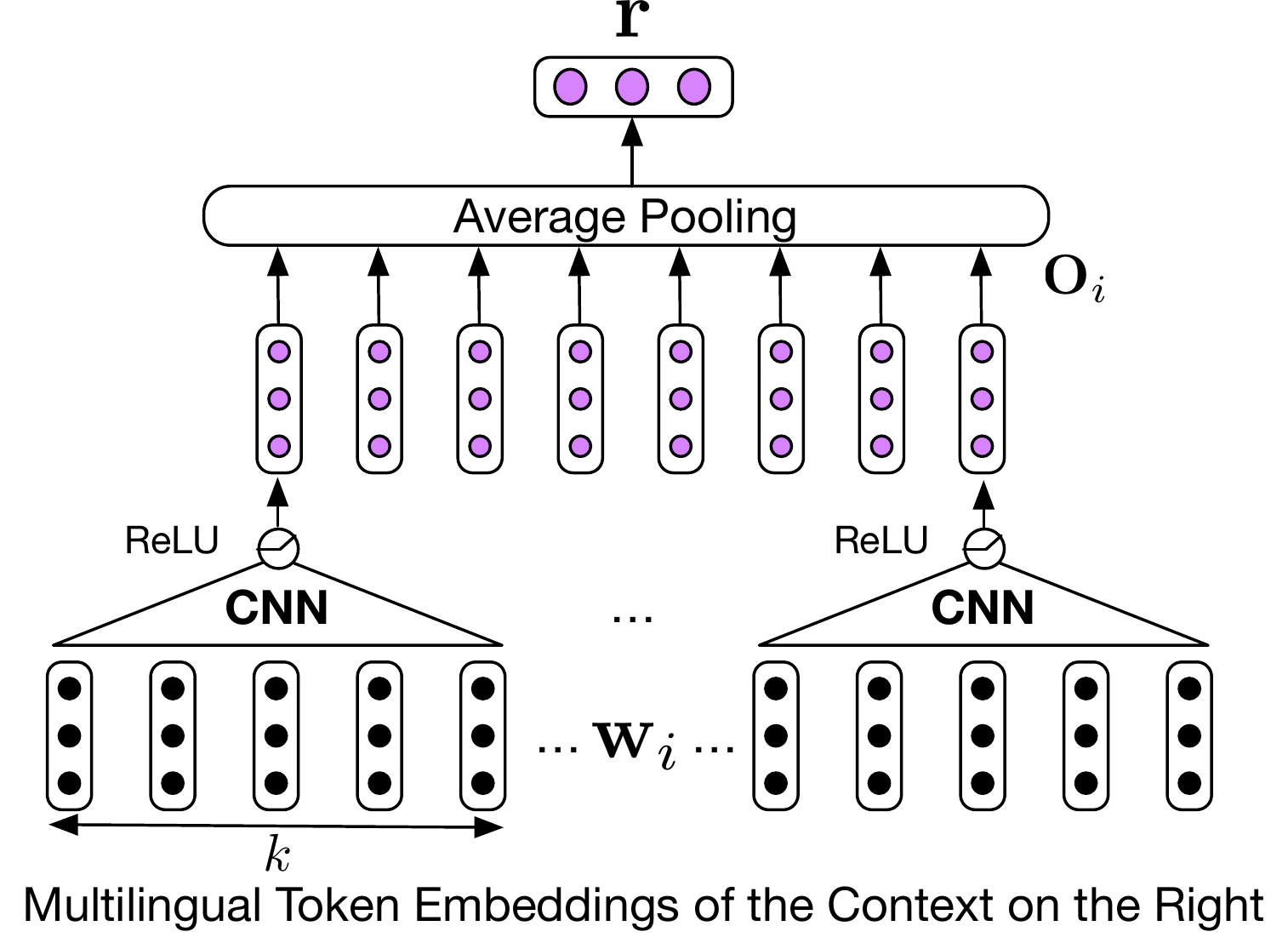}
  \caption{\footnotesize Local Context Encoder, for the right context. Figure~\ref{fig:mentionenc} shows how it fits inside Mention Context Encoder.}
  \label{fig:loccxt}
\end{figure}

\paragraph{Local Context Representation}
The local context of a mention $m$, spanning tokens
$i$ to $j$, consists of left context (tokens $i-W$ to $j$) and right context (tokens $i$ to $j+W$). For example, for the mention \ment{Liverpool} in Figure~\ref{fig:mentionenc}, the left and right contexts are ``Everton won against Liverpool'' and ``Liverpool in a FA Cup match'' respectively. The local context encoder (Figure~\ref{fig:loccxt}) encodes the left and the right contexts into vectors $\mathbf{l} \in \mathbb{R}^h$ and $\mathbf{r} \in \mathbb{R}^h$ using a convolutional neural network (CNN). These two vectors are then combined to generate the local context vector $\mathbf{c} \in \mathbb{R}^h$ (Figure~\ref{fig:mentionenc}).

The CNN convolves continuous spans of $k$ tokens using
a filter matrix $\mathbf{F} \in \mathbb{R}^{kd \times h}$ to project
the concatenation ($\oplus$ operator) of the token embeddings in the span. The resulting
vector is passed through a ReLU unit to
generate convolutional output $\mathbf{O}_i$. The outputs $\{\mathbf{O}_i\}$ are pooled by averaging,
\begin{align}
  & \mathbf{O}_i = \text{\sc relu}(\mathbf{F}^T (\mathbf{w}_i \oplus \cdots \oplus \mathbf{w}_{i+k-1})) \\
  & {\text{\sc Enc}}(\mathbf{w}_{1:n}) = \text{\sc avg}(\mathbf{O}_1,\cdots,\mathbf{O}_{n-k+1})
\end{align}
Left and right context vectors $\mathbf{l}$ and $\mathbf{r}$ are computed using respective ${\text{\sc Enc}}(.)$ layers, 
\begin{align}
  \mathbf{l} &= \text{\sc Enc}_{\text{left}}(\mathbf{w}_{i-W} \cdots \mathbf{w}_j) \\
  \mathbf{r} &= \text{\sc Enc}_{\text{right}}(\mathbf{w}_{i} \cdots \mathbf{w}_{j+W})
\end{align}
These vectors together generate the local context vector $\mathbf{c} = \feedfwd_{2h,h}(\mathbf{l} \oplus \mathbf{r})$. Here $\feedfwd_{d_i,d_o}: \mathbf{v}_i \rightarrow \mathbf{v}_o$ denotes a feed-forward layer that takes $\mathbf{v}_i \in \mathbb{R}^{d_i}$ as input, and outputs $\mathbf{v}_o \in \mathbb{R}^{d_o}$.

% Prelim exp with LSTM/GRU found poorer results that CNN in
% multilingual setting, suggesting relying on word order might be
% detrimental.
\paragraph{Document Context Representation}
Presence of certain mentions in a document can help disambiguate other mentions. For example, ``Suarez'', ``Everton'' in a document can help disambiguate ``Liverpool''. To incorporate this, we define the document context $d_m$ of a mention $m$ appearing in document $\doc$ to be the
bag of all other mentions in $\doc$.
We encode $d_m$ into a dense document context vector
$\mathbf{d}\in \mathbb{R}^h$ by a feed-forward layer $\mathbf{d} = \feedfwd_{|V|,h}(d_m)$.
Here $V$ is the set containing all mention surfaces seen during training. When training jointly over multiple languages, $V$ consists of mention surfaces seen in all
languages (e.g. all English and Tamil mention surfaces) during training. This enables parameter
sharing by embedding mention surfaces in different
languages in the same low-dimensional space.

\noindent The local and document context vectors $\mathbf{c}$ and $\mathbf{d}$ are combined to get the mention context vector $\mathbf{g} = \feedfwd_{2h,h}(\mathbf{c} \oplus \mathbf{d})$. 

\paragraph{Context Conditional Probability} 
We compute the probability of a mention $m$ linking to entity $e$ using its context vector $\mathbf{g}$ and the entity vector $\mathbf{e}$,
\begin{align}
  \text{P}_{\text{context}}(e \mid m) = \frac{\exp(\mathbf{g}^T\mathbf{e})}{\sum\limits_{e' \in C(m)}\exp(\mathbf{g}^T\mathbf{e'})} \label{eqn:cxtprob}
\end{align}
where $C(m)$ denotes all candidate entities of the mention $m$ (\S\ref{sec:candidate-generation} explains how $C(m)$ is generated). We minimize the negative log-likelihood of $\text{P}_{\text{context}}(e \mid m)$ with respect to the gold entity $e^{*}$ against the candidate entities $C(m)$, and call it the Entity-Context loss ({\sc EC-Loss}),
% \begin{align*}
%   \text{\sc EC-Loss} \!=\! - \log \text{P}_{\text{context}}(e^{*} \mid m) + \log \sum_{e' \in C(m)} \text{P}_{\text{context}}(e' \mid m)
% \end{align*}
\begin{align}
  \text{\sc EC-Loss} \!=\! - \log \frac{\text{P}_{\text{context}}(e^{*} \mid m)}{\sum\limits_{e' \in C(m)} \text{P}_{\text{context}}(e' \mid m)} \label{eqn:ecloss}
\end{align}
\subsection{Including Type Information}
\label{subsec:type}
Incorporating the fine-grained types of a mention $m$ can help rank entities of the appropriate type higher than others~\cite{ling2015design,Gupta17,Raiman2018}. For instance, knowing the correct type of mention \ment{Liverpool} as {\tt sports\_team} and constraining linking to entities with the relevant type, encourages disambiguation to the correct entity.

To make the mention context representation $\mathbf{g}$ type-aware, we predict the set of fine-grained types of $m$, $\mathbf{T}(m) =\{t_1,...,t_{|\mathbf{T}(m)|}\}$ using $\mathbf{g}$. Each $t_i$ belongs to a pre-defined type vocabulary $\Gamma$.\footnote{We use the type vocabulary $\Gamma$ from \newcite{LingWe12}, which contains 112 fine-grained types ($|\Gamma|=112$)} The probability of a type $t$ belonging to $\mathbf{T}(m)$ given the mention context is defined as
$\text{P}(t \mid m) \!=\!
\sigma(\mathbf{t}^T\mathbf{g})$, where $\sigma$ is the sigmoid function and $\mathbf{t}$ is the learnable embedding for type $t$.  

\noindent
We define a Type-Context loss ({\sc TC-Loss}) as,
\begin{align}
  \text{{\sc TC-Loss}}= \text{\sc BCE}(\mathbf{T}(m),\text{P}(t \mid m)) 
  \label{eqn:ctloss}
%   \text{{\sc TE-Loss}}= \text{\sc BCE}(\mathbf{T}(m),\text{P}(t \mid e^{*})) \label{eqn:etloss}
\end{align}
where BCE is the Binary Cross-Entropy Loss,
\begin{align*}
  % \begin{split}
    -\!\!\!\!\sum\limits_{t \in \mathbf{T}(m)} \!\! \log \text{P}(t \mid m)
    -\!\!\!\!\sum\limits_{t \not\in \mathbf{T}(m)} \!\! \log( 1- \text{P}(t \mid m))
  % \end{split}
\end{align*}
We also incorporate the entity-type information in the entity representations, and define a similar Type-Entity loss ({\sc TE-Loss}).

To identify the gold types $\mathbf{T}(m)$ of a mention $m$, we make the distant supervision assumption (same as \citet{ling2015design}) and assign the types of the gold entity $e^*$ to be the types of the mention.
Gold fine-grained types of the entities can be acquired from resources like Freebase~\cite{BEPST08} or YAGO~\cite{HSBW13}.

%%% Local Variables:
%%% mode: latex
%%% TeX-master: "main"
%%% End:

\section{Training and Inference}
We explain how \name~ generates candidate entities, performs inference, and combines the different training losses.

\subsection{Candidate Generation}
\label{sec:candidate-generation}
Candidate generation identifies a small number of plausible entities for a mention $m$ to avoid brute force comparison with all KB entities.
Given $m$, candidate generation outputs a
list of candidate entities $C(m)=\{e_1,e_2,\cdots,e_K\}$ of
size at most $K$ (we use $K$=20), each associated with a prior probability
$\text{P}_{\text{prior}}(e_i \mid m)$ indicating the probability of $m$ referring to $e_i$, given only $m$'s surface.
$\text{P}_{\text{prior}}$ is estimated from counts over the training mentions.

We adopt \newcite{TsaiRo16b}'s candidate generation strategy with some
minor modifications (Appendix A).
Using other approaches like CrossWikis~\cite{SPITKOVSKY12}, lead to consistently worse recall.
We note that transliteration based candidate
generation~\cite{mcnamee2011cross,P17-1178,TsaiRo18,UKR18} can further improve
recall. 

\subsection{Inference}
We combine the context conditional entity
probability $\text{P}_{\text{context}}(e \mid m)$ (eq.\ \ref{eqn:cxtprob}) and prior probability 
$\text{P}_{\text{prior}}(e \mid m)$ by taking their union:
\begin{align*}
  \begin{split}
  \text{P}_{\text{model}}(e \mid m) &= \text{P}_{\text{prior}}(e \mid m) + \text{P}_{\text{context}}(e \mid m) \\
  &- \text{P}_{\text{prior}}(e \mid m) \times \text{P}_{\text{context}}(e \mid m)
\end{split}
\end{align*}
Inference for the mention $m$ picks the entity,
% $\hat{e}= \text{argmax}_{C(m)} \text{P}_{\text{model}}(e \mid m)$.
\begin{align}
\hat{e}= \argmax_{e \in C(m)} \text{P}_{\text{model}}(e \mid m)
\end{align}

\subsection{Training Objective}
\label{sec:training}
When only training the mention context encoder and entity vectors, we minimize the {\sc EC-Loss}
averaged over all training mentions. When
using the two type-aware losses, we minimize a
weighted sum of {\sc EC-Loss}, {\sc TE-Loss}, and {\sc TC-Loss},
using the weighing scheme of \newcite{kendall2017multi},
\begin{align}
  \begin{split}
  & \frac{\text{\sc EC-Loss}}{2\lambda_{\text{\sc EC}}^2}  + \frac{\text{\sc TE-Loss}}{2\lambda_{\text{\sc TE}}^2}  + \frac{\text{\sc TC-Loss}}{2\lambda_{\text{\sc TC}}^2}  \\
  & + \log \lambda_{\text{\sc EC}}^2 + \log \lambda_{\text{\sc TE}}^2 + \log \lambda_{\text{\sc TC}}^2
  \end{split}
\end{align}
Here $\lambda_{i}$ are learnable scalar weighing parameters, and the respective $\frac{1}{2\lambda_{i}^2}$ and $\log\lambda_{i}^2$ term
ensure that $\lambda_{i}^2$ does not grow unboundedly. This way, the model learns the relative
weight for each loss term. 

During training, mentions from different languages are mixed using
{\em inverse-ratio mini-batch mixing} strategy. That is, if two
languages have training data sizes proportional to $\alpha:\beta$, at
any time during training, mini-batches seen from them are in the ratio
$\frac{1}{\alpha}:\frac{1}{\beta}$. This strategy prevents languages
with more training data from overwhelming languages with less training
data. Though simple, we found this strategy yielded good results.

%%% Local Variables:
%%% mode: latex
%%% TeX-master: "main"
%%% End:

\section{Experimental Setup}
\label{sec:experimental-setup}
We briefly describe the training and evaluation datasets, and the previous XEL approaches from the literature used in our comparison.
\subsection{Training Mentions}
\label{sec:training-data}
Following previous work, we use hyperlinks from Wikipedia (dumps dated 05/20/2017) 
as our source of grounded mentions for supervision. 
Wikipedias in different languages have different pages for the same entity, which are resolved by using inter-language links (e.g., page \begin{CJK*}{UTF8}{gbsn}利物浦\end{CJK*} in Chinese Wikipedia resolves to {\tt Liverpool} in English).  Training mentions statistics are shown in Table~\ref{tab:xlmotiv}.
\begin{table}[t]
  \footnotesize
  \centering
  \begin{tabular}{cC{2.5cm}C{2.5cm}}
    \toprule
    Lang. & \# Train Mentions & Size Relative to \# English Mentions\\
    \midrule
    German (de) & 22.6M & 43.7\%\\
    Spanish  (es) & 13.8M & 26.7\%\\
    French (fr) & 16.2M & 31.3\%\\
    Italian (it) & 11.5M & 22.2\%\\
    Chinese (zh) & 5.9M & 11.4\%\\
    Arabic (ar) & 3.1M & 6.0\%\\
    Turkish (tr) & 1.8M & 3.5\%\\
    Tamil (ta) & 473k & 0.9\%\\
    \bottomrule
  \end{tabular}
  \caption{\footnotesize Number of train mentions (from Wikipedia) in each
    language, with \% size relative to
    English (51.7M mentions). Train mentions from Wikipedias like Arabic, Turkish and Tamil are <10\% the size of those from the English Wikipedia.}
  \label{tab:xlmotiv}
\end{table}

We evaluate on 8 languages -- German (de), Spanish (es), Italian (it), French (fr),
Chinese (zh), Arabic (ar), Turkish (tr) and Tamil (ta), each of which has varying amount of grounded mentions from the respective Wikipedia (Table~\ref{tab:xlmotiv}).
We note that our method is applicable to any of the $293$ Wikipedia
languages as a target language.
% \ignore{The languages have Wikipedias of varying size, from {\sc Large} (German, Spanish, Italian, French) containing $>$10M linked
% mentions, {\sc Medium} (Chinese, Arabic) containing between 2M and 10M
% linked mentions, to {\sc Small} (Turkish and Tamil) with $<$2M linked
% mentions.}

\subsection{Evaluation Datasets}
We evaluate \name~ on the following benchmark datasets, spanning 8 different languages, thus providing an extensive evaluation.

\paragraph{McN-Test}
dataset from~\cite{mcnamee2011cross}. 
The test set was collected by using parallel document collections, and then crowd-sourcing the ground truths.
All the test mentions in this dataset consists of person-names only.

\paragraph{TH-Test}
A subset of the dataset used in~\cite{TsaiRo16b}, derived from Wikipedia.\footnote{\newcite{P17-1178} also created a dataset using Wikipedia, but did not categorize mentions like \newcite{TsaiRo16b}. Preliminary experiments on their dataset showed \name~ consistently beat \newcite{P17-1178}'s model. We chose {\sc TH-Test} for more controlled experiments.}
The mentions in the dataset fall in two categories -- {\em easy} and
{\em hard}, where hard mentions are those for which the most likely
candidate according to the prior probability (i.e., $\argmax \text{P}_{\text{prior}}(e \mid m)$) is {\em not} the correct
title.
Indeed, most Wikipedia mentions can be correctly linked by selecting the most likely candidate~\cite{RRDA11}.
We use all the hard mentions from \newcite{TsaiRo16b}'s test
splits for each language, and collectively call this
subset {\sc TH-Test}.

\paragraph{TAC15-Test}
TAC 2015~\cite{ji2015overview} dataset for Chinese and Spanish. 
It contains documents from discussion forum articles and news. \\

\noindent We evaluate all models using linking accuracy on gold mentions,
and assume gold mentions are provided at test time.
Table~\ref{tab:evaldata} summarizes the different domains of the evaluation datasets.

\begin{table}[h]
  \footnotesize
  \centering
  \begin{tabular}{lp{1.5cm}L{3cm}}
    Dataset & Lang. & Source\\
    \toprule
    \multirow{2}{*}{{\sc TH-Test}} & de, es, fr, it, & \multirow{2}{*}{Wikipedia} \\
    & zh, ar, tr, ta & \\
    \midrule
%     {\sc McN-Test}  & de, es, fr, it, zh, ar, tr & Parliamentary Proceedings (Europarl), News (SETimes, ProjSynd)\\
    \multirow{2}{*}{{\sc McN-Test}} & {de, es, fr, it,} & News,\\
    & zh, ar, tr & Parliament Proceedings\\
    \midrule
    \multirow{2}{*}{{\sc TAC15-Test}} & \multirow{2}{*}{es, zh} & News,\\
     &  & Discussion Forums\\
    \bottomrule
  \end{tabular}
  \caption{\footnotesize Evaluation datasets used in our experiments.}
  \label{tab:evaldata}
\end{table}

%%% Local Variables:
%%% mode: latex
%%% TeX-master: "main"
%%% TeX-PDF-mode: t
%%% End:

\paragraph{Tuning} 
We avoid any dataset-specific tuning, instead tuning on a development
set and applying the same parameters across all datasets. All tunable
parameters were tuned on a development set containing the hard
mentions from the train split released by \newcite{TsaiRo16b}. We
refer the reader to Appendix B for details on
tuning.

\subsection{Comparative Approaches}
\label{sec:comp-appr}
We compare against the following state-of-the-art (SoTA) approaches, described with the language from which they use mention contexts in {\bf (.)},

\paragraph{\newcite{TsaiRo16b} (Target Only)} trains a separate XEL model for each language using mention contexts from the target language Wikipedia only. Current SoTA on {\sc TH-Test}.

\paragraph{\newcite{P17-1178} (English Only)} 
uses entity coherence statistics from English Wikipedia and the document context of a mention for XEL. Current SoTA on {\sc McN-Test}, except for Italian and Turkish, for which it's \newcite{mcnamee2011cross}.

\paragraph{\newcite{Sil2018} (English Only)} uses multilingual embeddings to transfer a pre-trained English entity linking model to perform XEL for Spanish and Chinese. Prior probabilities $\text{P}_{\text{prior}}$ are used as a feature. Current SoTA on {\sc TAC15-Test}.

%%% Local Variables:
%%% mode: latex
%%% TeX-master: "main"
%%% TeX-PDF-mode: t
%%% End:

\section{Experiments}
\label{sec:exp}
We show that: {\bf (a)} \name~ can train a better entity linking model for a target language on various benchmark datasets by exploiting additional data from a high resource language like English (\S\ref{sec:main-experiments}).
{\bf (b)} \name~ can train a \emph{single} XEL model for multiple related languages and improve upon separately trained models (\S\ref{sec:mult-train}).
{\bf (c)} Adding additional type information as multi-task loss to \name~ further improves performance (\S\ref{sec:adding-type-inform}).

In all tables, we report the linking accuracy of \name~, averaged over 5 different runs, and mark with $^*$ the statistical significance ($p<0.01$) of the best result (shown {\bf bold})
against the state-of-the-art (SoTA) using Student's one-sample t-test.

\subsection{Monolingual and Joint Models}
\label{sec:main-experiments}
In Table~\ref{tab:maintable} and \ref{tab:tactable} we compare
\name~(mono), which uses monolingual supervision in the target
language only, and \name~(joint), which uses supervision from
English in addition to the monolingual supervision, with the state-of-the-art approaches.

We see that \name~(mono) achieves similar
or slightly better scores than respective SoTA on all datasets. The SoTA for {\sc McN-Test} in Turkish and Chinese enhances the model by using transliteration for candidate generation, explaining their superior performance.
\name~(joint) performs substantially better than
\name~(mono) on all datasets (Table~\ref{tab:maintable} and
\ref{tab:tactable}), proving that using additional supervision from a high resource language like 
English leads to better linking performance. In particular, \name~(joint) outperforms the SoTA on all languages in {\sc TH-Test}, on Spanish in TAC15-Test, and on 4 of the 7 languages in {\sc McN-Test}.

\begin{table}[t]
  \tabcolsep=1mm
  \footnotesize
  \centering
  % \begin{tabular}{cC{0.6cm}C{0.6cm}C{0.6cm}C{0.6cm}C{0.6cm}C{0.6cm}C{0.6cm}|C{0.6cm}C{0.6cm}C{0.6cm}C{0.6cm}C{0.6cm}C{0.6cm}C{0.6cm}}
  % \begin{tabular}{cccccccc|ccccccc}
  % \begin{tabular}{cp{0.6cm}p{0.6cm}p{0.6cm}p{0.6cm}p{0.6cm}p{0.6cm}}
  \begin{tabular}{c@{\hskip 2mm}c@{\hskip 2mm}c@{\hskip 2mm}c@{\hskip 6mm}c@{\hskip 2mm}c@{\hskip 2mm}c}
    \toprule
    {{\scriptsize \bf Dataset} {\tiny $\rightarrow$}} & \multicolumn{3}{c@{\hskip 6mm}}{{\sc TH-Test}} & \multicolumn{3}{c}{{\sc McN-Test}} \\
    % \midrule
    \cmidrule(l{0mm}r{6mm}){2-4}
    \cmidrule(l{0mm}r{0mm}){5-7}
    \multirow{2}{*}{{\scriptsize \bf Lang} {\tiny $\downarrow$}} & \multirow{2}{*}{SoTA} & \multicolumn{2}{c@{\hskip 6mm}}{\name} & \multirow{2}{*}{SoTA} & \multicolumn{2}{c@{\hskip 2mm}}{\name} \\
    &  & mono  & joint &  & mono  & joint \\
    % \midrule
%     \cmidrule(l{0mm}r{2mm}){2-2}
    \cmidrule(l{0mm}r{6mm}){2-4}
%     \cmidrule(l{0mm}r{2mm}){5-5}
    \cmidrule(l{0mm}r{0mm}){5-7}
    de & 53.3 & 53.7  & \bf 55.6$^*$    & 89.7 & 90.9  & \bf 91.5 \\
    % \midrule
    es & 54.5 & 54.9  & \bf 56.6$^*$    & {\bf 91.5} & 91.2 & 91.4 \\
    % \midrule
    fr & 47.5 & 48.5  & \bf 49.9$^*$  & 92.1 & 92.6  & \bf 92.7 \\
    % \midrule
    it & 48.3 & 48.4  & \bf 51.9$^*$  & 85.9 & 87.0  & \bf 87.8$^*$ \\
    % % \midrule
    zh & 57.6 & 58.1  & \bf 61.3$^*$  & {\bf 91.2}$^*$ & 87.4 & 88.2 \\
    % \midrule
    ar & 62.1 & 62.6  & \bf 63.8$^*$   & 80.2 & 80.3 & \bf 83.1$^*$ \\
    % \midrule
    tr & 60.2 & 61.0  & \bf 61.7$^*$   & {\bf 95.3}$^*$ & 91.0 & 91.9 \\
    % \midrule
    ta & 54.1 & 54.7  & \bf 59.7$^*$   & n/a & n/a & n/a \\
    \midrule
    avg. & 54.7 & 55.2 & \bf 57.6 & 89.4 & 88.6 & \bf 89.5 \\
    \bottomrule
  \end{tabular}
  \caption{\footnotesize \name~(joint) improves upon \name~(mono) and the current
    State-of-The-Art (SoTA) on {\sc TH-Test} and {\sc McN-Test}, showing the benefit of using additional supervision from English. The best score is shown {\bf bold} and $^*$ marks statistical significance of best against SoTA. Refer \S\ref{sec:comp-appr} for details on SoTA. 
% \tododr{DR: Maybe  makes sense to say something about the McN data set; as a minimum, it seems to be easy enough that you might be close to the ceiling.}
  }
  \label{tab:maintable}
  % \begin{tikzpicture}[overlay,remember picture]
  %   \draw[->] let \p1=(start), \p2=(end) in ($(\x1,\y1)-(0.5,-0.2)$) -- node[rotate=90,above] {decreasing size} ($(\x1,\y2)-(0.5,-0.2)$);
  % \end{tikzpicture}
\end{table}

%%% Local Variables:
%%% mode: latex
%%% TeX-master: "main"
%%% End:
\begin{table}[t]
  \footnotesize
  \centering
  \begin{tabular}{lccc}
    \toprule
    \multicolumn{2}{c}{{\scriptsize \bf Model} {\tiny $\downarrow$} {\scriptsize \bf Lang.} {\tiny $\rightarrow$}} & es & zh \\
    \midrule
    \multicolumn{2}{c}{\cite{TsaiRo16b}} & 82.4 & 85.1\\
    \multicolumn{2}{c}{\cite{Sil2018} (SoTA)} & 83.9 & 85.9\\
    \midrule
    \parbox[t]{2mm}{\multirow{7}{*}{\rotatebox[origin=c]{90}{\name~}}} & mono & 83.3 & 84.4 \\
                                               & mono$^{\text{+type}}$ & 83.5 & 84.8 \\
    \cmidrule(lr){2-2}
                                               % & Joint - prior &  &  \\
                                               & joint & 84.1 & 85.5 \\
                                               & joint$^{\text{+type}}$ & \bf 84.4$^*$ & \bf 86.0 \\
    \cmidrule(lr){2-2}
                                               % & Multi - prior &  &  \\
                                               & multi & 83.9 & n/a\\
                                               & multi$^{\text{+type}}$ & {\bf 84.4}$^*$ & n/a\\
    % \cmidrule(lr){2-2}
    %                                            & ZS - prior & {\bf 84.3} & --\\
    %                                            & ZS & {\bf 84.3} & --\\
    \bottomrule
  \end{tabular}
  \caption{\footnotesize Linking accuracy on TAC15-Test. Numbers for \newcite{Sil2018} from personal communication.}
  \label{tab:tactable}
\end{table}

%%% Local Variables:
%%% mode: latex
%%% TeX-master: "main"
%%% End:
% The magnitude of the improvement on {\sc TH-Test} varies with
% language, with average improvement for high resource languages like
% Spanish, German, Italian and French being approx 2.1\%, and varying
% from 0.7 to 5.0\% for other languages. Overall, the joint training
% improves performance by 1.9\% over \name~(Mono) on {\sc TH-Test},
% averaged across all languages.
%
%
\subsection{Multilingual Training}
\label{sec:mult-train}
\begin{table}[t]
  \tabcolsep=1mm
  \footnotesize
  \centering
  % \begin{tabular}{cp{0.6cm}p{0.6cm}p{0.6cm}p{0.6cm}p{0.6cm}p{0.6cm}}
  \begin{tabular}{c@{\hskip 2mm}c@{\hskip 2mm}c@{\hskip 2mm}c@{\hskip 6mm}c@{\hskip 2mm}c@{\hskip 2mm}c}
    \toprule
    {{\scriptsize \bf Dataset} {\tiny $\rightarrow$}} & \multicolumn{3}{c@{\hskip 6mm}}{{\sc TH-Test}} & \multicolumn{3}{c}{{\sc McN-Test}} \\
    % \midrule
    \cmidrule(l{0mm}r{6mm}){2-4}
    \cmidrule(l{0mm}r{0mm}){5-7}
    \multirow{2}{*}{{\scriptsize \bf Lang} {\tiny $\downarrow$}} & \multirow{2}{*}{SoTA} & \multicolumn{2}{c@{\hskip 6mm}}{\name} & \multirow{2}{*}{SoTA} & \multicolumn{2}{c@{\hskip 2mm}}{\name} \\
     &  & joint & multi &  & joint  & multi \\
    \cmidrule(l{0mm}r{6mm}){2-4}
    \cmidrule(l{0mm}r{0mm}){5-7}
    de & 53.3 & \bf 55.6$^{*}$  &  55.2   & 89.7 & \bf 91.5 & 91.4 \\
    % \midrule
    es & 54.5 & 56.6  &  \bf 56.8$^{*}$   & \bf 91.5 & 91.4 & 91.4 \\
    % \midrule
    fr & 47.5 & 49.9  &  \bf 51.0$^{*}$  & 92.1 & \bf 92.7  & 92.6 \\
    it & 48.3 & 51.9  &  \bf 52.3$^{*}$  & 85.9 & 87.8  & \bf 87.9$^{*}$ \\
    \midrule
    avg. & 50.9 & 53.5 & \bf 53.8 & 89.8 & \bf 90.8 & \bf 90.8  \\
    \bottomrule
  \end{tabular}
  \caption{\footnotesize Linking accuracy of a {\em single} \name~(multi) model for four languages -- German, Spanish, French and Italian. Individually trained \name~(joint) scores are also shown. The best score is shown {\bf bold} and $^*$ marks statistical significance of {\bf best} against SoTA. Refer \S\ref{sec:comp-appr} for details on SoTA.}
  \label{tab:multitable}
\end{table}

%%% Local Variables:
%%% mode: latex
%%% TeX-master: "main"
%%% End:
\name~ is the first approach that can train a \emph{single} XEL model for multiple languages.
To demonstrate this capability, we train a model, henceforth referred as \name~(multi), 
{\em jointly} on 5 related languages -- Spanish, German, French, Italian and
English. We compare \name~(multi) to the respective \name~(joint) model for each language.

Table~\ref{tab:tactable} and \ref{tab:multitable}, show that \name~(multi) is better (or at par) than \name~(joint) on all datasets. 
This shows that \name~(multi) can making more efficient use of available supervision in related languages than previous approaches which trained separate models per language.

\subsection{Adding Fine-grained Type Information}
\label{sec:adding-type-inform}
To study the effect of adding fine-grained type information, in Table~\ref{tab:tactable} we compare \name~(mono) and \name~(joint) to \name~(mono$^{\text{+type}}$) and \name~(joint$^{\text{+type}}$) respectively, which are versions of \name~(mono) and \name~(joint) trained using the two type-aware losses. 

\name~(mono$^{\text{+type}}$) and \name~(joint$^{\text{+type}}$) both improve compared to \name~(mono) and \name~(joint) on {\sc McN-Test} and {\sc
  TH-Test} (Table~\ref{tab:typetable} vs Table~\ref{tab:maintable}), showing the benefit of using
structured knowledge in the form of fine-grained types. Similar trends are also seen on {\sc TAC15-Test}
(Table~\ref{tab:tactable}), where \name~(joint$^{\text{+type}}$)
improves on the SoTA for Spanish and Chinese.
\begin{table}[h]
  \tabcolsep=1mm
  \footnotesize
  \centering
  % \begin{tabular}{cp{0.6cm}p{0.6cm}p{0.6cm}p{0.6cm}p{0.6cm}p{0.6cm}}
  \begin{tabular}{c@{\hskip 0mm}c@{\hskip 1mm}c@{\hskip 1mm}c@{\hskip 3mm}c@{\hskip 1mm}c@{\hskip 1mm}c}
    \toprule
    {{\scriptsize \bf Dataset} {\tiny $\rightarrow$}} & \multicolumn{3}{c@{\hskip 3mm}}{{\sc TH-Test}} & \multicolumn{3}{c}{{\sc McN-Test}} \\
    % \midrule
    \cmidrule(l{0mm}r{3mm}){2-4}
    \cmidrule(l{0mm}r{0mm}){5-7}
    % \multirow{2}{*}{{\scriptsize \bf Lang} {\tiny $\downarrow$}} & \multirow{2}{*}{SoTA} & \multicolumn{2}{c@{\hskip 3mm}}{\name~+ type} & \multirow{2}{*}{SoTA} & \multicolumn{2}{c@{\hskip 1mm}}{\name~+ type} \\
    \multirow{2}{*}{{\scriptsize \bf Lang} {\tiny $\downarrow$}} & \multirow{2}{*}{SoTA} & \multicolumn{2}{c@{\hskip 3mm}}{\name} & \multirow{2}{*}{SoTA} & \multicolumn{2}{c@{\hskip 1mm}}{\name} \\
     &  & mono$^{\text{+type}}$ & joint$^{\text{+type}}$ &  & mono$^{\text{+type}}$ & joint$^{\text{+type}}$ \\
%     \cmidrule(l{0mm}r{1mm}){2-2}
    \cmidrule(l{0mm}r{3mm}){2-4}
%     \cmidrule(l{0mm}r{1mm}){5-5}
    \cmidrule(l{0mm}r{0mm}){5-7}
    de & 53.3 & 54.0  & \bf 55.9$^*$  & 89.7 & 91.2  &  \bf 91.5 \\
    % \midrule
    es & 54.5 & 55.1  & \bf 57.2$^*$  & \bf 91.5 & 91.0 & 91.2 \\
    % \midrule
    fr & 47.5 & 49.0  & \bf 50.6$^*$  & 92.1 & 92.6 & \bf 92.7 \\
    % \midrule
    it & 48.3 & 49.2  & \bf 52.2$^*$  & 85.9 & 87.4  & \bf 87.9$^*$ \\
    % % \midrule
    zh & 57.6 & 58.9  & \bf 61.5$^*$  & \bf 91.2$^*$ & 87.6 & 88.4\\
    % \midrule
    ar & 62.1 & 63.0  & \bf 64.0$^*$  & 80.2 & 81.1 & \bf 84.0$^*$ \\
    % \midrule
    tr & 60.2 & 61.5  & \bf 62.0$^*$  & \bf 95.3$^*$ & 91.2 &  92.1 \\
    % \midrule
    ta & 54.1 & 56.0  & \bf 59.9$^*$  & n/a & n/a & n/a \\
    \midrule
    avg. & 54.7 & 55.8 & \bf 57.9 & 89.4 & 88.9 & \bf 89.7 \\
    \bottomrule
  \end{tabular}
  \caption{\footnotesize Adding fine-grained type information further improves linking accuracy (compare to Table~\ref{tab:maintable}). The best score is shown {\bf bold} and $^*$ marks statistical significance of {best} against SoTA. Refer \S\ref{sec:comp-appr} for details on SoTA.}
  \label{tab:typetable}
\end{table}

%%% Local Variables:
%%% mode: latex
%%% TeX-master: "main"
%%% End:

%%% Local Variables:
%%% mode: latex
%%% TeX-master: "main"
%%% End:

\section{Experiments with Limited Resources}
\label{sec:low-reso-exper}
The key motivation of \name~ is to exploit supervision from high-resource languages like English to aid XEL for languages with limited resources. In this section, we examine two such scenarios,

\noindent {\bf (a)} {\em Zero-shot setting} i.e., no supervision available in the target language. 
Our analysis reveals the limitations of zero-shot XEL approaches and finds that the prior probabilities play an important role in achieving good performance (\S\ref{sec:zero-shot-setting}), which are unavailable in realistic zero-shot scenarios.

\noindent {\bf (b)} {\em Low-resource setting} i.e., some supervision available in the target language. We show that by combining supervision from a high-resource language, like English, \name~ can achieve competitive performance with a fraction of available supervision in the target language (\S\ref{sec:how-much-target}).

\subsection{Zero-shot Setting}
\label{sec:zero-shot-setting}
We first explain how \name~ can perform zero-shot XEL, the implications of our zero-shot setting, and how it is more realistic than previous work.

\paragraph{Zero-shot XEL with XELMS} \name~ performs zero-shot XEL by training a model using English supervision and multilingual embeddings for English, and directly applying it to the test data in another language using the respective multilingual word embedding instead of English embeddings. 

\paragraph{No Prior Probabilities}
Prior probabilities (or prior), i.e., $\text{P}_{\text{prior}}$ have been shown to be a reliable indicator of the correct disambiguation in entity linking ~\cite{RRDA11,TsaiRo16b}.
These probabilities are estimated from counts over the training mentions in the target language. In the absence of training data for the target language, as in the zero-shot setting, these prior probabilities are not available to an XEL model.

\paragraph{Comparison to Previous Work}
The only other model capable of zero-shot XEL is that of
\newcite{Sil2018}. However, \newcite{Sil2018} use prior probabilities and coreference chains for the target language in their zero-shot
experiments, both of which will not be available in a realistic zero-shot scenario. Compared to \newcite{Sil2018}, we evaluate the performance of zero-shot XEL
in more realistic setting, and show it is adversely affected by absence of prior probabilities.

\begin{table}[t!]
  \tabcolsep=1mm
  \footnotesize
  \centering
  \begin{tabular}{llcccc}
    \toprule
    \multicolumn{2}{c}{{{\scriptsize \bf Dataset} {\tiny $\rightarrow$}}} & \multicolumn{2}{c}{TAC15-Test} & TH-Test & McN-Test\\
    \multicolumn{2}{c}{{\scriptsize \bf Approach} {\tiny $\downarrow$}} & (es) & (zh) & (avg) & (avg) \\
     \midrule
    %(uses prior)
    %\multicolumn{2}{l}{\newcite{Sil2018}} & 83.9 & 85.9 & n/a & n/a \\
     %\midrule
    \multicolumn{2}{l}{\name~ (Z-S w/ prior)} & 80.3 & 83.9 & 43.5 & 88.1\\
    \multicolumn{2}{l}{\name~ (Z-S w/o prior)} & 53.5 & 55.9 & 41.1 & 86.0\\
     \midrule
     %& SoTA & 83.9 & 85.9 & 54.7 & 89.4\\
     \multicolumn{2}{l}{SoTA} & 83.9 & 85.9 & 54.7 & 89.4\\
                                               
    % \cmidrule(lr){2-2}
    %                                            & joint w/o prior & 77.3 & 79.4 & 57.0 & 88.5\\
    %                                            & joint w/ prior & 84.1 & 85.5 & 57.6 & 89.5 \\
    % \cmidrule(lr){2-2}
    %                                            & multi w/o prior & 79.5 & -- \\
    %                                            & multi w/ prior &  & -- \\
%     \midrule
%     \multicolumn{2}{c}{prior only} & 78.2 & 79.8 & & \\
    \bottomrule
  \end{tabular}
  \caption{\footnotesize Linking accuracy of the zero-shot (Z-S) approach on different datasets. Zero-shot (w/ prior) is close to SoTA for datasets like TAC15-Test, but performance drops in the more realistic setting of zero-shot (w/o prior) (\S\ref{sec:zero-shot-setting}) on all datasets, indicating most of the performance can be attributed to the presence of prior probabilities. 
  The slight drop in {\sc McN-Test} is due to trivial mentions, which only have a single candidate.
  }
  \label{tab:zsnoprior}
\end{table}
% SoTA for McN-Test and TH-Test are from Table~\ref{tab:typetable} and \ref{tab:maintable} \todo{should I show w/o type SoTA?}
% \ignore{\todo{OPTION 2: distinguish zero shot models from non-zs in
%     the table, isntead of currnet layout.}}
%     In contrast, Joint (w/o prior) achieves better scores, comparable to scores of Joint (w/ prior), showing the Joint approach does not suffer in the absence of priors.
%     Scores for Joint (w/ prior) are from Table~\ref{tab:maintable} and \ref{tab:tactable}. 
%     McN-Test has little ambiguity (see text for details).\todo{Detailed scores for TH-Test and McN-Test are in Appendix.}
%     \ignore{This is contrary to findings \todo{should I add:on {\sc TAC15-Test}?} in previous zero-shot XEL approaches~\cite{Sil2018}}

\paragraph{Is zero-shot XEL really effective?} 
To evaluate the effectiveness of the zero-shot XEL approach, we perform zero-shot XEL using \name~ on all datasets. Table~\ref{tab:zsnoprior} shows zero-shot XEL results on all datasets, both with and without using the prior during inference. Note that zero-shot XEL (with prior) is close to SoTA (\newcite{Sil2018}) on {\sc TAC15-Test}, which also uses the prior for zero-shot XEL. However, for zero-shot XEL (without prior) performance drops by more than 20\% for TAC15-Test, 2.4\% for TH-Test and by 2.1\% for McN-Test. This indicates that zero-shot XEL is not effective in a realistic zero-shot setting (i.e., when the prior is unavailable for inference).

We found that the prior is indeed a strong indicator of the correct disambiguation. For instance, simply selecting the the most likely candidate using the prior for {\sc TAC15-Test} achieved 77.2\% and 78.8\% for Spanish and Chinese respectively. 
It is interesting to note that both zero-shot XEL (with or without prior) perform worse than the best possible model on {\sc TH-Test}, because {\sc TH-Test} was constructed to ensure prior probabilities are not strong indicators~\cite{TsaiRo16b}.
On {\sc McN-Test}, we found that an average of 75.9\% mentions have only one (the correct) candidate, making them trivial to link, regardless of the absence of priors.

The results show that most of the XEL performance in zero-shot settings can be attributed to availability of prior probabilities for the candidates.
It is evident that zero-shot XEL in a realistic setting (i.e., when prior probabilities are not available) is still a challenging problem.

\subsection{Low-resource Setting}
\label{sec:how-much-target}
We analyze the behavior of \name~ in a low-resource setting, i.e. when some supervision is available in the target language.
The aim of this setting is to estimate how much supervision from the target language is needed to get reasonable performance when using it jointly with supervision from English.
To discount the effect of prior probabilities, we report all results without the prior.

\begin{figure}[t]
  \centering
  \begin{tikzpicture}[scale=0.9]
    \begin{axis}[
      xlabel=\# train mentions in target language (in millions),
      y label style={at={(axis description cs:0.05,.5)}},
%       legend style={draw=none}, % no border around legend
      ylabel=Linking Accuracy,
      xtick={0,0.25,0.50,0.75,1.0,1.25,1.50},
      % title=Varying \# of Target Lang. Mention Contexts,
      minor y tick num=9,
      ymax=63,
      ymin=36,
      xmin=0.0,
      xmax=1.0,
      legend style={
        at={(0.5,1.27)},       % 0.5 times xmin,xmax and -0.18 times ymin,ymax, think affine comb.
        anchor=north,
        legend columns=3,       % 2 columns
        /tikz/every even column/.append style={column sep=0.6cm}
      },
      ]
      \addplot[mark=None,very thick,dotted,color=blue] table [x=frac, y=tr-mono, col sep=comma] {data_frac_results.txt};
      \addplot[mark=triangle,very thick,color=blue] table [x=frac, y=tr-joint, col sep=comma] {data_frac_results.txt};
      \addplot[mark=None,very thick,dashed,color=blue] table [x=frac, y=tr-best, col sep=comma] {data_frac_results.txt};

      \addplot[mark=None,very thick,dotted,color=red] table [x=frac, y=zh-mono, col sep=comma] {data_frac_results.txt};
      \addplot[mark=o,very thick,color=red] table [x=frac, y=zh-joint, col sep=comma] {data_frac_results.txt};
      \addplot[mark=None,very thick,dashed,color=red] table [x=frac, y=zh-best, col sep=comma] {data_frac_results.txt};

      \addplot[mark=None,very thick,dotted,color=green!60!black] table [x=frac, y=es-mono, col sep=comma] {data_frac_results.txt};
      \addplot[mark=square,very thick,color=green!60!black] table [x=frac, y=es-joint, col sep=comma] {data_frac_results.txt};
      \addplot[mark=None,very thick,dashed,color=green!60!black] table [x=frac, y=es-best, col sep=comma] {data_frac_results.txt};

      % \legend{Tr-Best, Tr-Joint, Zh-Best, Zh-Joint, Es-Best, Es-Joint}
      % \legend{Tr-Mono, Tr-Joint, Zh-Mono, Zh-Joint, Es-Mono, Es-Joint}
      \legend{
      \footnotesize tr-mono, 
        \footnotesize tr-joint, 
        \footnotesize tr-best, 
        \footnotesize zh-mono, 
        \footnotesize zh-joint, 
        \footnotesize zh-best, 
        \footnotesize es-mono, 
        \footnotesize es-joint,
        \footnotesize es-best}

    \end{axis}
  \end{tikzpicture}
  \caption{\footnotesize Linking accuracy vs. the number of train mentions in the target language L (= Turkish (tr), Chinese (zh) and Spanish (es)). We compare both \name~(mono) and \name~(joint) to the best results using all available supervision, denoted by L-best. To discount the effect of the prior, all results above are without it. For number of train mentions = 0, \name~(joint) is equivalent to zero-shot without prior. Best viewed in color.
    % Observe that adding even a small fraction of mention contexts from the target language for Turkish and Chinese leads to scores comparable to a Joint model with full target language supervision.
    % Observe that \name~(Joint) significantly outperforms \name~(Mono) in the low data regime ($<$0.5M), and given a small number of training contexts, can consistently achieve higher accuracies than \name~(Mono).
  }
  \label{fig:datafrac}
\end{figure}
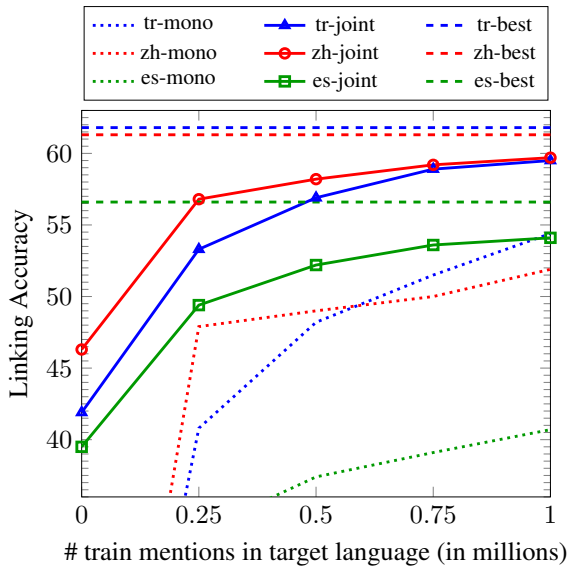

Figure~\ref{fig:datafrac} plots results on the TH-Test dataset when training a \name~(joint) model by gradually increasing the number of mention contexts for target language L (= Spanish, Chinese and Turkish) that are available for supervision. Figure~\ref{fig:datafrac} also shows the best results achieved using all available target language supervision (denoted by L-best). For comparison with the mono-lingually supervised model, we also plot the performance of \name~(mono), which only uses the target language supervision.

Figure~\ref{fig:datafrac} shows that after training on 0.75M mentions from Turkish and Chinese (and 1.0M mentions from Spanish), the \name~(joint) model is within 2-3\% of the respective L-best model which uses all training mentions in the target language, indicating that \name~(joint) can reach competitive performance even with a fraction of the full target language supervision. For comparison, a \name~(mono) model trained on the same number of training mentions is 5-10\% behind the respective \name~(joint) model, showing better utilization of target language supervision by \name~(joint).
% \todo{can we give a explanation for why different frac of supervision is needed?}

\ignore{Figure~\ref{fig:datafrac} shows that \name~(joint) consistently
outperforms the \name~(mono), with the gain being quite large in the
low data regime (0 to 0.5M). This confirms the value of training
mentions in the target language, as even a small amount helps
dramatically. The gap becomes smaller as more mentions are seen,
nevertheless \name~(joint) is ahead by at least 3\% even with 1.5M
training mentions, re-enforcing the benefit of joint training.
}

%%% Local Variables:
%%% mode: latex
%%% TeX-master: "main"
%%% End:

\section{Related Work}
Existing approaches have taken two main directions to obtain supervision for learning XEL models
--- {\bf (a)} using mention contexts appearing in the target language~\cite{mcnamee2011cross,TsaiRo16b}, or {\bf (b)} using mention contexts appearing only in English \cite{P17-1178, Sil2018}. We describe these directions and their limitations below, and explain how \name~ overcomes these limitations.

\newcite{mcnamee2011cross} use annotation projection via parallel corpora to generate mention contexts in the target
language, while \newcite{TsaiRo16b} learns separate XEL
models for each language and only use mention contexts in the target language.
Both these approach have scalability issues
for languages with limited resources.
Another limitation of these
approaches is that they train separate models for each language, which is inefficient when working with multiple languages.
\name~ overcomes these limitations as it can use mention context from multiple languages simultaneously, and train a single model. 

Other approaches only use mention contexts from English. While \newcite{P17-1178} compute entity coherence statistics from English Wikipedia, \newcite{Sil2018} perform zero-shot XEL for Chinese and Spanish by using multilingual embeddings to transfer a pre-trained English EL model. 
However, our work suggests that mention contexts in the target language should also be used, if available. Indeed, a recent study~\cite{lewoniewski2017relative} found that for
language sensitive topics, the quality of information can be
better in the relevant language version of Wikipedia than the English version.
Our analysis also shows that zero-shot XEL approaches like that of \newcite{Sil2018} are not effective in realistic zero-shot scenarios where good prior probabilities are unlikely to be available. 
In such cases, we showed that combining supervision available in the target language with supervision from a high-resource language like English can yield significant performance improvements.

The architecture of \name~ is inspired by several monolingual entity linking
systems~\cite{francislandau-durrett-klein:2016:N16-1,nguyen-EtAl:2016:COLING,Gupta17}, approaches that use type information to aid entity linking~\cite{ling2015design,Gupta17, Raiman2018}, and the recent success of multilingual embeddings for several tasks~\cite{TACL892,Duong-EtAl:2017:EACLlong}.

%%% Local Variables:
%%% mode: latex
%%% TeX-master: "main"
%%% TeX-PDF-mode: t
%%% End:

% \section{Analysis}
% \label{sec:analysis}
% \input{analysis}

\section{Conclusion}
We introduced \name~, an approach that can combine supervision from multiple languages to train an XEL model. We illustrate its benefits through extensive evaluation on different benchmarks. \name~ is also the first approach that can train a \emph{single} model for multiple languages, making more efficient use of available supervision than previous approaches which trained separate models.

Our analysis sheds light on the poor performance of zero-shot XEL in realistic scenarios where the prior probabilities for candidates are unlikely to exist, in contrast to findings in previous work that focused on high-resource languages. We also show how in low-resource settings, \name~ makes it possible to achieve competitive performance even when only a fraction of the available supervision in the target language is provided.

Several future research directions remain open. For all XEL approaches, the
task of candidate generation is currently limited by existence of a
target language Wikipedia and remains a key
challenge. A joint inference framework which enforces coherent
predictions~\cite{ChengRo13,globerson-EtAl:2016:P16-1,ganea-hofmann:2017:EMNLP2017}
could also lead to further improvements for XEL. Similar techniques
can be applied to other information extraction tasks like relation extraction to extend
them to multilingual settings.

%%% Local Variables:
%%% mode: latex
%%% TeX-master: "main"
%%% End:

\section*{Acknowledgments}
The authors would like to thank Snigdha Chaturvedi, Anne Cocos, Stephen Mayhew, Chen-Tse Tsai, Qiang Ning, Jordan Kodner, Dan Deutsch, John Hewitt and the anonymous EMNLP reviewers for their useful comments and suggestions. 

This work was supported by Contract HR0011-15-2-0025 and Agreement HR0011-15-2-0023 with the US Defense Advanced Research Projects Agency (DARPA). Approved for Public Release, Distribution Unlimited. The views expressed are those of the authors and do not reflect the official policy or position of the Department of Defense or the U.S. Government.

% \bibliography{emnlp2018}
\bibliography{references,ccg,cited}
\bibliographystyle{acl_natbib_nourl}

% \appendix
% \input{appendix}

\end{document}